\documentclass{article}

 \usepackage[preprint]{neurips_2026}

\usepackage[utf8]{inputenc} 
\usepackage[T1]{fontenc}    
\usepackage{hyperref}       
\usepackage{url}            
\usepackage{booktabs}       
\usepackage{amsfonts}       
\usepackage{nicefrac}       
\usepackage{microtype}      
\usepackage{xcolor}

\usepackage[T1]{fontenc}
\usepackage{newtxtext}
\title{Reward-Gated On-Policy Distillation}

\usepackage{amsmath, amssymb, amsthm, amsfonts}
\usepackage{bm}
\usepackage{microtype}
\usepackage{booktabs}
\usepackage{algorithm}
\usepackage{algpseudocode}
\usepackage{hyperref}
\usepackage{amssymb}
\usepackage{bbm}

\usepackage[utf8]{inputenc} 
\usepackage[T1]{fontenc}    
\usepackage{hyperref}       
\usepackage{url}            
\usepackage{booktabs}       
\usepackage{amsfonts}       
\usepackage{nicefrac}       
\usepackage{microtype}      
\usepackage{xcolor}         
\usepackage{graphicx}
\usepackage[inline]{enumitem}

\usepackage{pgfplots}
\pgfplotsset{compat=1.18}
\usepackage{xcolor}

\usepackage{pgfplots}

\usepackage[table]{xcolor}
\usepackage[most]{tcolorbox}

\pgfplotsset{compat=1.18}
\usetikzlibrary{matrix}

\definecolor{newblue}{HTML}{0064E0}

\definecolor{nblue}{HTML}{E3EDFF}

\definecolor{ngreen}{HTML}{F3F1E8}

\definecolor{nred}{HTML}{FFE8ED}

\newcommand{\topK}{\mathrm{top}_K}

\definecolor{topkcolor}{RGB}{31, 119, 180}   
\definecolor{tailcolor}{RGB}{214, 90,  0}    
\definecolor{green_pos}{HTML}{6A994E}        
\definecolor{red_neg}{HTML}{C44536}          

\hypersetup{
  colorlinks,
  linkcolor=newblue,
  citecolor=newblue,
  urlcolor=newblue
}

\newtcolorbox[auto counter]{mainbox}[2][]{%
  colframe=ngreen, 
  colback=ngreen,  
  breakable,
  before upper={{\textbf{Main Contributions.} #2}},
  #1
}

\hypersetup{
  colorlinks,
  linkcolor=newblue,
  citecolor=newblue,
  urlcolor=newblue
}

\newcommand{\given}{\,\vert\,}
\newcommand{\indicator}{\mathbbm{1}}
\newcommand{\KL}{\mathrm{KL}}
\newcommand{\stud}{\pi_\theta}
\newcommand{\studold}{\pi_{\theta_{\mathrm{old}}}}
\newcommand{\teach}{\pi_T}
\newcommand{\Lkl}{\mathcal{L}_{\mathrm{KL}}}
\newcommand{\Lrt}{\mathcal{L}_{\mathrm{RG\text{-}OPD}}}

\usepackage{xspace}
\newcommand{\methodname}{\textsc{RG-OPD}\xspace}

\author{%
  Mohammad Sadegh Akhondzadeh\\
   University of Cologne\\
\href{ms.akhondzadeh@gmail.com}{\texttt{ms.akhondzadeh@gmail.com
}} 
  \And
  Vijay Lingam\\
  AWS AI\\
  \href{mailto:vijaylingam0810@gmail.com}{\texttt{vijaylingam0810@gmail.com}}
  \\
  \AND
  Atula Tejaswi\\
  UT Austin\\
  \And
  Chanakya Ekbote
  \\
  MIT
  \\
  \AND
  Sujay Sanghavi \\
   UT Austin\\
  \And
  Aleksandar Bojchevski \\
   University of Cologne 
}

\usepackage[inline]{enumitem}

\begin{document}

\maketitle


\begin{abstract}
On-policy distillation is a powerful way to transfer reasoning ability from a
strong teacher to a smaller student: the student samples trajectories from its
own policy, and the teacher provides dense token-level supervision on the states
the student actually visits. However, this supervision is not always reliable: a teacher can assign high likelihood to plausible but incorrect
solutions, or low likelihood to correct student solutions that follow different
reasoning paths. Unconditionally distilling the teacher can therefore reinforce
bad modes or erase useful student behavior. To address these limitations, we introduce \methodname{}: \textbf{R}eward-\textbf{G}ated \textbf{O}n-\textbf{P}olicy \textbf{D}istillation that uses verifier feedback to decide when teacher logits should be trusted. \methodname{} bridges sparse verifier rewards and dense teacher logits,
preserving token-level supervision while filtering misleading teacher signals.
Across reasoning and coding benchmarks, RG-OPD produces stronger distilled
students, outperforming both vanilla reverse-KL distillation and the recent
TSD-KD baseline. At 1K generation length, RG-OPD improves over reverse-KL by
\textbf{2.9} points and over TSD-KD by \textbf{4.9} points; in the
long-generation setting, it improves over the untuned student by \textbf{8.2}
points. Our code is available at \url{https://github.com/UoC-tail/RG-OPD}.
\end{abstract}


\section{Introduction}
\label{sec:introduction}

On-policy distillation (OPD) is an effective way to transfer reasoning ability from a strong teacher model to a smaller student. In OPD, the student samples trajectories from its own policy, and the teacher provides dense token-level supervision on the states the student actually visits~\citep{gkd}. This makes OPD especially attractive for reasoning post-training: verifier rewards from RLVR provide only sparse trajectory-level feedback, while teacher logits provide a learning signal at every token. In its standard form, OPD applies this supervision unconditionally, treating the teacher as a reliable oracle for all rollouts. However, a teacher may assign high likelihood to a plausible but incorrect solution, or low likelihood to a correct student solution that follows a different reasoning path. Unconditional distillation can therefore reinforce incorrect modes or suppress useful student behaviors.

Based on these observations, we introduce \methodname{}: \textbf{R}eward-\textbf{G}ated \textbf{O}n-\textbf{P}olicy \textbf{D}istillation, which addresses the following question: \emph{\textbf{When does teacher supervision help rather than hurt during on-policy distillation?}} At each training step, the student samples trajectories on policy, a verifier assigns each trajectory a scalar reward or
advantage, and a fixed teacher provides token-level distributions over the
student's sampled states. Instead of applying the distillation loss to every
trajectory, \methodname{} keeps only trajectories for which the verifier reward
and the teacher--student likelihood gap agree. For reward-positive trajectories,
we distill only when the teacher assigns higher likelihood than the student; for
reward-negative trajectories, we distill only when the teacher assigns lower
likelihood than the student.

This simple gate preserves the main advantage of logit-level distillation: dense token-level supervision from a stronger model. At the same time, it avoids the central failure mode of unconditional distillation: treating the teacher as reliable even when its likelihoods conflict with observed correctness. Thus, \methodname{} provides a direct bridge between sparse verifier rewards and dense teacher logits. Unlike RLVR~\citep{shao2024deepseekmath, dr_grpo, ekbote2025murphy}, it does not rely solely on terminal rewards; unlike standard on-policy distillation~\citep{gu2024minillm, gkd}, it does not assume the teacher is always
correct; and unlike joint RL--KD objectives~\citep{xu2025kdrl, hubotter2026sdpo, zhang2026rlad}, it explicitly conditions each
distillation update on trajectory-level reliability.

\begin{mainbox}{} \textbf{(i)} We identify a reliability mismatch in on-policy distillation for reasoning: teacher likelihoods and verifier rewards can disagree on student-generated trajectories, making unconditional distillation harmful. \textbf{(ii)} We introduce RG-OPD, a reward-gated likelihood filter that distills from the teacher only when the verifier reward and teacher--student likelihood gap are directionally aligned. \textbf{(iii)} Empirically, RG-OPD produces stronger distilled students, outperforming vanilla reverse-KL distillation by \textbf{2.9} points at 1K generation length and \textbf{2.8} points at 8K, while improving over the untuned student by \textbf{8.2} points in the long-generation setting.
\end{mainbox}

\section{Related Work}
\label{sec:related_work}

Reinforcement learning with verifiable rewards (RLVR) has become a practical
post-training paradigm for reasoning tasks with automatically scored outputs
\citep{shao2024deepseekmath, guo2025deepseekr1}. PPO, GRPO, and related variants
optimize student-generated rollouts from sparse outcome rewards
\citep{schulman2017ppo, shao2024deepseekmath, gspo, dr_grpo,
ekbote2025murphy}, but terminal feedback provides limited token-level credit
assignment, especially when correct samples are rare. Knowledge distillation
offers a complementary signal by transferring dense token-level supervision from
a stronger teacher~\citep{hinton_distillation}; on-policy distillation further
reduces exposure bias by querying the teacher on student-generated trajectories.
GKD formalizes on-policy distillation for autoregressive language models,
querying the teacher on student-generated outputs with divergences such as
reverse KL or Jensen--Shannon divergence~\citep{gkd}. TSD-KD builds on this
student-centric perspective with preference-based and token-selective
distillation~\citep{tsd_kd}. However, TSD-KD applies teacher
supervision broadly and does not test whether the teacher is reliable on a given
trajectory. Recent and concurrent work also combines verifier rewards with
teacher-like supervision: KDRL adds teacher reverse-KL regularization to
GRPO-style optimization~\citep{xu2025kdrl}; RL-aware distillation anchors the
policy ratio to a mixture of the old student and teacher policies
\citep{zhang2026rlad}; Self-Distilled RLVR uses a privileged variant of the same
policy for token-level guidance~\citep{yang2026selfdistilledrlvr}; and related
self-distillation methods derive dense guidance from privileged or
feedback-conditioned student variants~\citep{zhao2026opsd, hubotter2026sdpo}.
These methods show that dense teacher-like signals can improve reward-based
reasoning post-training, but they either apply distillation as a broad auxiliary
signal, modify the RL objective, or focus on self-distillation. In contrast,
RG-OPD studies external-teacher on-policy distillation and gates each
trajectory-level distillation update based on agreement between verifier reward
and the teacher--student likelihood gap.

\section{Reward-Gated On-Policy Distillation}
\label{sec:method}
 
\textbf{On-policy distillation.} We consider on-policy distillation of a student policy $\stud$ from a fixed teacher $\teach$. At each iteration, for each prompt $x$, the student samples $N$ trajectories from its own policy: $\{y^{(i)}\}_{i=1}^{N} \sim \stud(\cdot \given x)$, and a verifier returns a scalar reward $r_i$ for each trajectory. For every sampled token $y_t^{(i)}$ we have access to three log-probabilities evaluated at the \emph{sampled action}: the current student $\log \stud(y_t^{(i)} \given s_t^{(i)})$, the behavior policy at sampling time $\log \studold(y_t^{(i)} \given s_t^{(i)})$, and the teacher $\log \teach(y_t^{(i)} \given s_t^{(i)})$. We follow \citep{hubotter2026sdpo, sdft} and apply the top-k reverse-KL (Appendix~\ref{app:topk_kl}) distillation loss i.e.
$\Lkl^{(i)} \;=\; \sum_t m_t^{(i)} \,\KL\!\left(\pi_\theta(\cdot \given s_t^{(i)}) \,\Vert\, \pi_T(\cdot \given s_t^{(i)})\right)$, 
where $m_t^{(i)}$ is the response mask. This loss is applied uniformly across sampled trajectories, regardless of whether the trajectory was successful and regardless of whether the teacher actually provides useful information at it.
 
\textbf{When the teacher and the reward disagree.} In the RLVR context, vanilla on-policy distillation can lead to two failure modes:
\begin{enumerate*}[label=(\roman*)]
\item Teacher disagrees with a student success: the student samples a correct continuation that the teacher would have been unlikely to produce. Pulling the student toward the teacher in this case \emph{unlearns} a behavior the reward signal certifies as good. \item Teacher endorses a student failure: The student samples a wrong trajectory to which the teacher also assigns high probability. Reverse-KL has nothing to push against here, the teacher's mode coincides with the bad mode.
\end{enumerate*}
A reward signal alone resolves neither case (it cannot tell the student \emph{what} to do differently), and a teacher signal alone resolves neither (it has no information about correctness on a freshly sampled trajectory). The resolution is to use them jointly: the reward decides \emph{whether} the trajectory is good, and the teacher--student likelihood gap decides \emph{whether the teacher is informative} about the cause.\looseness=-1
 
\begin{figure}
    \centering
    \includegraphics[width=1\linewidth]{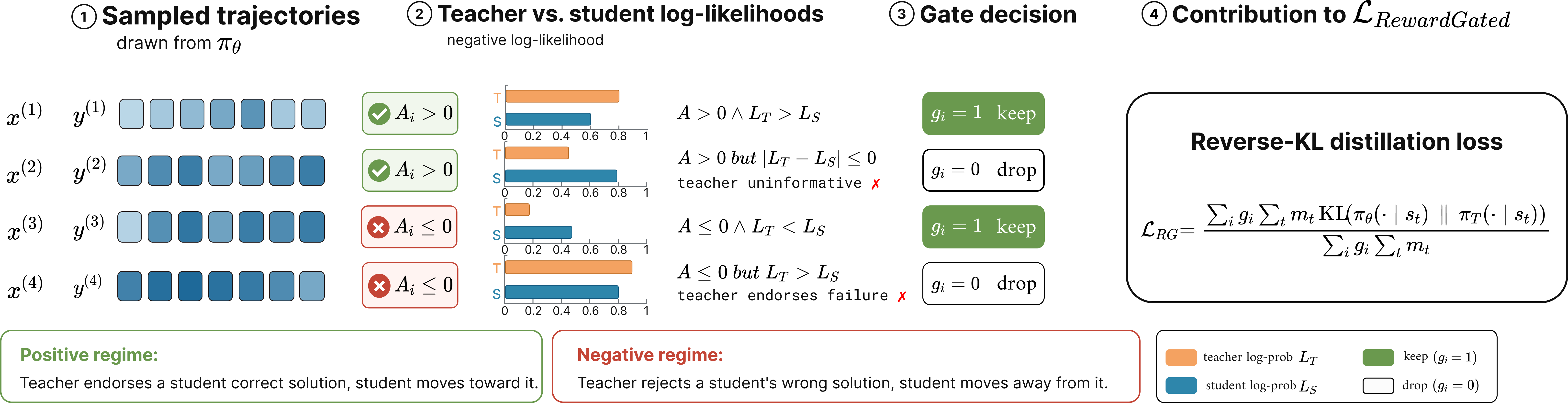}
    \caption{Overview of \methodname{}. The student samples trajectories and receives verifier rewards $A_i$. The reward--teacher gate keeps a trajectory ($g_i{=}1$) only when the teacher--student likelihood gap is directionally consistent with the reward: $L_T > L_S$ for correct trajectories and $L_T < L_S$ for incorrect ones. The reverse-KL distillation loss is computed exclusively over kept trajectories.}
    \label{fig: digram}
\end{figure}

\textbf{Reward-Teacher Likelihood Gate.} For each sampled trajectory $i$, define the trajectory-level teacher and student log-likelihoods evaluated on the sampled tokens,
\begin{equation}
L_T^{(i)} \;=\; \sum_t m_t^{(i)} \log \teach\!\left(y_t^{(i)} \given s_t^{(i)}\right), \qquad
L_S^{(i)} \;=\; \sum_t m_t^{(i)} \log \stud\!\left(y_t^{(i)} \given s_t^{(i)}\right),
\label{eq:LTLS}
\end{equation}
and let $A_i$ denote the trajectory-level reward signal (we use the GRPO advantage of $y^{(i)}$ in our experiments). We define the \emph{reward--teacher gate}
\begin{equation}
g_i \;=\; \indicator\!\left[\,\textcolor{green_pos}{\big(A_i > 0 \,\wedge\, L_T^{(i)} > L_S^{(i)} + \delta\big)} \;\vee\; \textcolor{red_neg}{\big(A_i \leq 0 \,\wedge\, L_T^{(i)} < L_S^{(i)} - \delta\big)}\,\right],
\label{eq:gate}
\end{equation}
with an optional confidence margin $\delta \geq 0$ (default $\delta = 0$). The gated distillation loss is
\begin{equation}
\Lrt \;=\; \frac{\sum_i g_i \sum_t m_t^{(i)} \, \KL\!\left(\stud(\cdot \given s_t^{(i)}) \,\Vert\, \teach(\cdot \given s_t^{(i)})\right)}{\sum_i g_i \sum_t m_t^{(i)}}.
\label{eq:gated-loss}
\end{equation}
The student is updated only on trajectories for which the teacher gives a directionally useful signal relative to the observed reward; all other trajectories are dropped from the distillation loss. The same reverse-KL objective is used in both kept regimes; only the set of trajectories contributing to the loss changes.\looseness=-1


\section{Experiments}
Following~\citep{tsd_kd}, we use Qwen2.5-1.5B/14B-\textsc{Instruct} models as the student-teacher pair and train on a subset of the \textsc{UltraInteract} \citep{yuan2024advancing} for 3 epochs. We provide implementation details in Appendix~\ref{app:implementation}.

\textbf{Results.} In \autoref{tab:kd_main_shortcontext}, we evaluate all methods with the default maximum generation length of 1024 tokens.
\methodname{} achieves the best average score (46.94) among all distilled models, outperforming the vanilla Reverse-KL baseline (44.02) by 2.9 points and TSD-KD (42.03).
Gains are consistent across math and coding tasks. An exception is IFEval, where TSD-KD achieves the best score (54.68) and \methodname{} scores 48.60. We attribute this to TSD-KD's student-centric design: rather than enforcing the teacher distribution uniformly, it combines preference-based indirect feedback, token-selective distillation, and entropy regularization, which may better preserve the student's instruction-following behavior while still providing teacher guidance.


\definecolor{deltagreen}{HTML}{1A9850}
\definecolor{deltared}{HTML}{D73027}
\definecolor{deltagray}{HTML}{777777}
\definecolor{oursblue}{HTML}{E8F1FF}

\newcommand{\deltag}[2]{%
    {\color{deltagreen}{\scriptsize$\uparrow$}~\textcolor{deltagreen}{+#1}}%
}

\newcommand{\deltar}[2]{%
    {\color{deltared}{\scriptsize$\downarrow$}~\textcolor{deltared}{-#1}}%
}

\newcommand{\deltaz}{%
    {\textcolor{deltagray}{0.00}}%
}

\newcommand{\oursrow}{%
    \rowcolor{oursblue}%
}

\begin{table*}[t]
    \centering
    \vspace{-2mm}
    \caption{
    Performance of knowledge distillation methods on generation-based benchmarks under
    1K and 8K maximum generation lengths. $\Delta$ is relative to the corresponding
    \textbf{${Qwen2.5(1.5B)_S}$} student baseline within each generation-length block
    (green/red = gain/drop). \textbf{Avg.\ Rank}$\downarrow$ is the mean rank across
    benchmarks among student and distillation methods, excluding the teacher reference.
    TSD-KD$^\ast$ numbers are borrowed from~\citep{tsd_kd} and are shown for reference
    only; they are excluded from Avg.\ Rank.
    }
    \label{tab:kd_main_shortcontext}
    \footnotesize
    \setlength{\tabcolsep}{3pt}
    \renewcommand{\arraystretch}{1.15}
    \resizebox{1\textwidth}{!}{%
    \begin{tabular}{l cc cc cc cc cc cc c}
        \toprule
        \textbf{Method}
        & \textbf{GSM8K} & \textbf{$\Delta$}
        & \textbf{GSM+} & \textbf{$\Delta$}
        & \textbf{MATH} & \textbf{$\Delta$}
        & \textbf{MPMath} & \textbf{$\Delta$}
        & \textbf{MBPP} & \textbf{$\Delta$}
        & \textbf{IFEval} & \textbf{$\Delta$}
        & \textbf{Avg.\ Rank$\downarrow$} \\
        \midrule
        \multicolumn{13}{c}{\textit{Generation Length 1024}} \\
        \midrule
        \textit{${Qwen2.5(14B)_T}$}
        & \textit{$76.83_{\pm 0.61}$} & --
        & \textit{$57.18_{\pm 0.41}$} & --
        & \textit{$30.33_{\pm 0.81}$} & --
        & \textit{$78.86_{\pm 0.79}$} & --
        & \textit{$53.13_{\pm 3.11}$} & --
        & \textit{$84.17_{\pm 1.02}$} & --
        & -- \\
        ${Qwen2.5(1.5B)_{S}}$
        & $56.89_{\pm 1.92}$ & \deltaz
        & $38.69_{\pm 1.09}$ & \deltaz
        & $27.47_{\pm 2.09}$ & \deltaz
        & $37.33_{\pm 1.37}$ & \deltaz
        & $43.73_{\pm 1.86}$ & \deltaz
        & $52.12_{\pm 1.64}$ & \deltaz
        & 2.7 \\
        \midrule
        Reverse KL~\citep{gkd}
        & $65.91_{\pm 1.15}$ & \deltag{9.02}{75}
        & $42.97_{\pm 0.80}$ & \deltag{4.28}{45}
        & $25.08_{\pm 1.10}$ & \deltar{2.39}{24}
        & $37.26_{\pm 0.93}$ & \deltar{0.07}{18}
        & $46.60_{\pm 0.60}$ & \deltag{2.87}{35}
        & $46.28_{\pm 1.50}$ & \deltar{5.84}{55}
        & 3.0 \\
        TSD-KD~\citep{tsd_kd}
        & $54.21_{\pm 0.55}$ & \deltar{2.68}{28}
        & $35.10_{\pm 0.61}$ & \deltar{3.59}{38}
        & $27.10_{\pm 0.58}$ & \deltar{0.37}{18}
        & $38.91_{\pm 1.27}$ & \deltag{1.58}{25}
        & $42.20_{\pm 1.04}$ & \deltar{1.53}{23}
        & $54.68_{\pm 0.98}$ & \deltag{2.56}{32}
        & 3.0 \\
        TSD-KD~\citep{tsd_kd}$^\ast$
        & $60.10$ & \deltag{3.21}{40}
        & $40.50$ & \deltag{1.81}{30}
        & $26.10$ & \deltar{1.37}{25}
        & $36.90$ & \deltar{0.43}{18}
        & $42.10$ & \deltar{1.63}{25}
        & $\mathbf{55.20}$ & \deltag{3.08}{40}
        & -- \\
        \oursrow
        \textbf{RG-OPD (ours)}
        & $\mathbf{66.69}_{\pm 0.56}$ & \deltag{9.80}{80}
        & $\mathbf{45.47}_{\pm 0.31}$ & \deltag{6.78}{60}
        & $\mathbf{28.63}_{\pm 1.27}$ & \deltag{1.16}{25}
        & $\mathbf{42.91}_{\pm 0.83}$ & \deltag{5.58}{55}
        & $\mathbf{49.33}_{\pm 0.76}$ & \deltag{5.60}{55}
        & $48.60_{\pm 0.87}$ & \deltar{3.52}{38}
        & \textbf{1.3} \\
        \midrule
        \multicolumn{13}{c}{\textit{Generation Length 8096}} \\
        \midrule
        \textit{${Qwen2.5(14B)_{T}}$}
        & \textit{$83.88_{\pm 0.31}$} & --
        & \textit{$67.44_{\pm 0.23}$} & --
        & \textit{$76.12_{\pm 0.18}$} & --
        & \textit{$79.77_{\pm 0.59}$} & --
        & \textit{$72.47_{\pm 2.53}$} & --
        & \textit{$84.53_{\pm 0.43}$} & --
        & -- \\
        ${Qwen2.5(1.5B)_{S}}$
        & $58.20_{\pm 2.14}$ & \deltaz
        & $40.63_{\pm 0.78}$ & \deltaz
        & $41.09_{\pm 3.51}$ & \deltaz
        & $37.38_{\pm 1.41}$ & \deltaz
        & $44.00_{\pm 1.93}$ & \deltaz
        & $52.24_{\pm 1.76}$ & \deltaz
        & 2.8 \\
        \midrule
        Reverse KL~\citep{gkd}
        & $72.63_{\pm 1.45}$ & \deltag{14.43}{80}
        & $52.33_{\pm 0.59}$ & \deltag{11.70}{75}
        & $49.80_{\pm 0.83}$ & \deltag{8.71}{65}
        & $37.28_{\pm 0.96}$ & \deltar{0.10}{18}
        & $47.27_{\pm 0.81}$ & \deltag{3.27}{35}
        & $46.08_{\pm 1.87}$ & \deltar{6.16}{55}
        & 2.7 \\
        TSD-KD~\citep{tsd_kd}
        & $54.59_{\pm 0.62}$ & \deltar{3.61}{38}
        & $35.35_{\pm 0.56}$ & \deltar{5.28}{50}
        & $27.89_{\pm 0.65}$ & \deltar{13.20}{80}
        & $38.91_{\pm 1.27}$ & \deltag{1.53}{25}
        & $42.33_{\pm 0.99}$ & \deltar{1.67}{25}
        & $\mathbf{54.80}_{\pm 0.55}$ & \deltag{2.56}{32}
        & 3.2 \\
        \oursrow
        \textbf{RG-OPD (ours)}
        & $\mathbf{73.77}_{\pm 0.80}$ & \deltag{15.57}{80}
        & $\mathbf{54.50}_{\pm 0.19}$ & \deltag{13.87}{80}
        & $\mathbf{52.60}_{\pm 0.61}$ & \deltag{11.51}{75}
        & $\mathbf{42.88}_{\pm 0.86}$ & \deltag{5.50}{55}
        & $\mathbf{50.00}_{\pm 1.04}$ & \deltag{6.00}{60}
        & $48.68_{\pm 0.84}$ & \deltar{3.56}{38}
        & \textbf{1.3} \\
        \bottomrule
    \end{tabular}%
    }
\end{table*}

\begin{figure}[h!]
    \centering
    \input{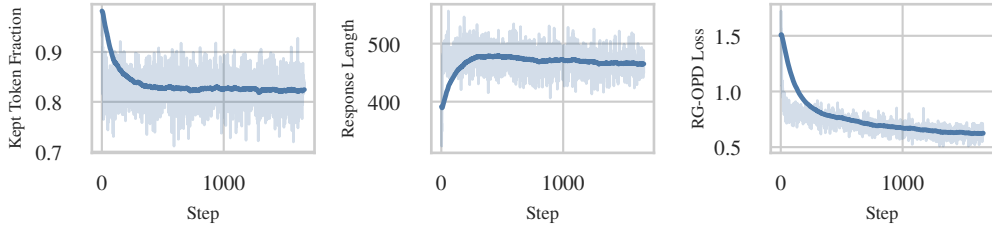}
    \caption{ \textbf{Training dynamics of \methodname{}.} \emph{Left}: the fraction of tokens retained by the reward--teacher gate decreases over training. \emph{Center}: mean response length grows throughout training. \emph{Right}: the gated reverse-KL distillation loss decreases over training.}
    \label{fig:training_dynamic}
\end{figure}

As shown in \autoref{fig:training_dynamic} [Center], distilled models tend to become more verbose over training, generating longer chains of thought on average.
We therefore repeat the evaluation with a generous generation budget of 8096 tokens.
With more room to reason, \methodname{} improves its average to 53.74, a 6.8-point gain over long-context Reverse-KL (50.90) and an 8.2-point improvement over the untuned student (45.59).
The gains are especially pronounced on MATH, where \methodname{} reaches 52.60 vs.\ 49.80. 

\textbf{Training Dynamics.}~\autoref{fig:training_dynamic} tracks three quantities over the course of \methodname{} training.
\emph{Kept Token Fraction} (left): the fraction of tokens admitted by the gate decreases over training, from roughly 0.9 to 0.7. This indicates that the teacher--student likelihood gap gradually shrinks, and fewer trajectories satisfy the alignment condition that triggers the gate. The gate therefore becomes more selective as training progresses, naturally reducing the distillation signal on trajectories where the teacher has little remaining information to offer.
\emph{Response Length} (center): the average number of generated tokens grows steadily throughout training; a consequence of imitating a teacher whose reasoning traces are more verbose.
\emph{\methodname{} Loss} (right): the gated reverse-KL loss decreases, confirming that the student is approaching the teacher's distribution on the gated trajectories.

\section{Conclusion}
We present \methodname, an on-policy distillation method that conditions teacher supervision on the agreement between verifier feedback and the teacher--student likelihood gap. The key insight is that dense teacher logits and sparse verifier rewards answer different questions. The reward certifies \emph{whether} a trajectory is good, while the likelihood gap reveals \emph{whether the teacher is directionally informative for the student}. By applying distillation only when these two signals are aligned, \methodname{} avoids the two failure modes of unconditional distillation: \begin{enumerate*}[label=(\roman*)]
\item reinforcing bad modes that the teacher endorses, \item erasing correct student behaviors that the teacher would not have produced. 
\end{enumerate*}
Experimental results show that RG-OPD outperforms both reverse-KL distillation and the TSD-KD baseline. 
Our results support the view that the value of a teacher signal on any given trajectory depends on the reward, and that a simple scalar gate is sufficient to exploit this structure.



\bibliographystyle{abbrv}
\bibliography{references}

\newpage
\appendix


\section{Implementation Details}
\label{app:implementation}
\subsection{Setup} 
We adopt the setup from \citep{tsd_kd}; we distill a \textsc{Qwen2.5-1.5B-Instruct} student from a frozen \textsc{Qwen2.5-14B-Instruct} teacher. The student is trained end-to-end with FSDP across 2 GPUs; the teacher is served via a dedicated vLLM endpoint (tensor-parallel $= 2$). At each training step, the teacher receives the student's rollouts and returns the top-$K$ log-probabilities at every position for KL distillation. Our training data is a subset of the \textsc{UltraInteract} \citep{yuan2024advancing}. We train for 3 epochs with AdamW (lr~$= 5\!\times\!10^{-6}$, cosine schedule, 50-step warmup). Top-$K$ logit distillation uses $K=50$ with tail correction (~\autoref{eq:topk-kl}) as default. Our evaluations are performed on lm-eval-harness\footnote{\url{https://github.com/EleutherAI/lm-evaluation-harness}}, following \citep{tsd_kd}.\looseness=-1

Finally for the evaluation, we evaluate on 10 benchmarks spanning math (GSM8K~\citep{gsm8k}, GSM-Plus~\citep{gsmplus}, MATH~\citep{math}, MMLU-Pro-Math~\citep{mmlupro} denoted as MPMath in our tables), code (MBPP~\citep{mbpp}), knowledge (SciQ~\citep{sciq}, MMLU-STEM~\citep{mmlupro}), reasoning (BBH~\citep{bbh}, MuSR~\citep{musr}), and instruction-following (IFEval~\citep{ifeval}), using lm-evaluation-harness with vLLM as the inference backend. All models are evaluated with temperature=0.6 over 3 seeds.

\begin{table}[h]
\centering
\caption{Ablation on the truncation parameter $k$ for top-$k$ knowledge distillation. The \emph{no-tail} variant omits the tail-redistribution step.}
\label{tab:topk_ablation}
\setlength{\tabcolsep}{4.5pt}
\renewcommand{\arraystretch}{1.15}
\resizebox{\textwidth}{!}{%
\begin{tabular}{l cccccc c}
\toprule
 & \multicolumn{7}{c}{\textbf{Generation-based}} \\
\cmidrule(lr){2-8}
\textbf{Method} & GSM8K & GSM+ & MATH & MPMath & MBPP & IFEval & \textbf{Avg.} \\
\midrule
Top-$k$ ($k{=}20$, no-tail)
 & $67.53_{\pm 0.50}$ & $45.45_{\pm 0.12}$ & $27.49_{\pm 1.34}$ & $38.44_{\pm 1.07}$ & $46.80_{\pm 3.03}$ & $47.28_{\pm 0.48}$ & 45.50 \\
Top-$k$ ($k{=}20$)
 & $67.58_{\pm 0.76}$ & $45.60_{\pm 0.57}$ & $28.33_{\pm 1.64}$ & $42.36_{\pm 0.87}$ & $47.67_{\pm 1.53}$ & $47.36_{\pm 0.67}$ & 46.48 \\
Top-$k$ ($k{=}30$)
 & $\mathbf{67.93}_{\pm 0.80}$ & $\mathbf{46.06}_{\pm 0.24}$ & $28.57_{\pm 1.12}$ & $\mathbf{43.38}_{\pm 1.19}$ & $49.27_{\pm 0.76}$ & $47.36_{\pm 0.43}$ & \textbf{47.10} \\
Top-$k$ ($k{=}50$)
 & $66.69_{\pm 0.56}$ & $45.47_{\pm 0.31}$ & $\mathbf{28.63}_{\pm 1.27}$ & $42.91_{\pm 0.83}$ & $\mathbf{49.33}_{\pm 0.76}$ & $\mathbf{48.60}_{\pm 0.87}$ & 46.94 \\
\bottomrule
\end{tabular}}
\end{table}
\subsection{Top-$K$ Reverse-KL Approximation.}
\label{app:topk_kl}
Computing $\KL(\stud\|\teach)$ naively requires storing the full vocabulary logits of both the student and the teacher simultaneously. To avoid this, similar to \citep{hubotter2026sdpo, sdft}, we use only the top-$K$ teacher logits indexed by the student's top-$K$ tokens and approximate the full-vocabulary reverse-KL as a \textcolor{topkcolor}{top-$K$ explicit sum} plus a \textcolor{tailcolor}{tail correction} that aggregates the remaining probability mass into a single bucket:
\begin{align}
&\KL\!\left(\stud(\cdot \given s_t^{(i)}) \,\Vert\, \teach(\cdot \given s_t^{(i)})\right) \notag \\
&\approx \textcolor{topkcolor}{\sum_{\hat{y}_t \,\in\, \topK(\stud)}
  \stud\!\left(\hat{y}_t \given s_t^{(i)}\right)
  \cdot \log \frac{\stud\!\left(\hat{y}_t \given s_t^{(i)}\right)}{\teach\!\left(\hat{y}_t \given s_t^{(i)}\right)}} \notag \\
&\quad + \textcolor{tailcolor}{%
    \Bigl(1 - \!\!\!\sum_{\hat{y}_t \,\in\, \topK(\stud)}\!\!\!
      \stud\!\left(\hat{y}_t \given s_t^{(i)}\right)\Bigr)
    \cdot \log
    \frac{%
      1 - \sum_{\hat{y}_t \,\in\, \topK(\stud)} \stud\!\left(\hat{y}_t \given s_t^{(i)}\right)%
    }{%
      1 - \sum_{\hat{y}_t \,\in\, \topK(\stud)} \teach\!\left(\hat{y}_t \given s_t^{(i)}\right)%
    }
  },
\label{eq:topk-kl}
\end{align}
where $\topK(\stud)$ denotes the $K$ tokens with highest probability under the current student $\stud$.  The \textcolor{topkcolor}{blue term} accumulates the per-token KL contribution over the top-$K$ support and carries the primary gradient signal for the student. The \textcolor{tailcolor}{orange tail term} handles the residual mass: it collapses all out-of-top-$K$ tokens into a single aggregate token then forms a binary KL between the student's and teacher's residual probabilities. The tail term also remains differentiable and gradients flow through it during backpropagation. \autoref{tab:topk_ablation} ablates the effect of $K$ and the tail correction. Removing the \textcolor{tailcolor}{tail term} (i.e., dropping the residual-mass bucket) degrades performance, confirming that the tail gradient is a non-trivial part of the learning signal even though its contribution to the total KL is small. Varying $K$ shows diminishing returns beyond a moderate value, indicating that the student's probability mass is already well concentrated in the top-$K$ tokens.

\section{Additional Results}

\paragraph{Likelihood-Based Tasks}
We evaluate on tasks scored by log-likelihood ranking (SciQ, MMLU-STEM, MuSR, BBH), where the model assigns log-probabilities to a fixed set of answer choices and the highest-scoring option is selected as the prediction, following~\citep{tsd_kd}. We note, however, that log-likelihood ranking is not the most natural evaluation protocol for instruction fine-tuned models.
As shown in \autoref{tab:kd_gen_vs_lik}, \methodname{} does not consistently improve over the baseline under this protocol.
These models are optimized to generate free-form responses and are not calibrated for pairwise answer scoring.
This is evident from the teacher itself: despite being a 14B model that scores 80.01 on generation-based benchmarks, it achieves only 63.77 under likelihood-based evaluation and on SciQ, the 1.5B student baseline actually outscores the teacher (89.50 vs.\ 86.80).
Finally, when we evaluate the models through text generation and extract the final answers, we observe that \methodname{} again outperforms the previous methods.

\begin{table*}[h]
    \centering
    \vspace{0mm}
    \caption{Comparison of knowledge distillation methods under generation-based and likelihood-based evaluation on the same four benchmarks. $\Delta$ is relative to the student baseline \textbf{${Qwen2.5(1.5B)_S}$} (green/red = gain/drop), and \textbf{Avg.\ Rank}$\downarrow$ is the mean rank across the eight benchmark metrics among student and distillation methods, excluding the teacher reference.}
    \label{tab:kd_gen_vs_lik}
    \footnotesize
    \setlength{\tabcolsep}{1.8pt}
    \renewcommand{\arraystretch}{1.15}
    \resizebox{1\textwidth}{!}{%
    \begin{tabular}{l cc cc cc cc @{\hskip 8pt} cc cc cc cc c}
        \toprule
        & \multicolumn{8}{c}{\textbf{Generation-based}}
        & \multicolumn{8}{c}{\textbf{Likelihood-based}}
        & \textbf{Avg.\ Rank$\downarrow$} \\
        \cmidrule(lr){2-9} \cmidrule(lr){10-17}
        \textbf{Method}
        & \textbf{SciQ} & \textbf{$\Delta$}
        & \textbf{MSTEM} & \textbf{$\Delta$}
        & \textbf{MuSR} & \textbf{$\Delta$}
        & \textbf{BBH} & \textbf{$\Delta$}
        & \textbf{SciQ} & \textbf{$\Delta$}
        & \textbf{MSTEM} & \textbf{$\Delta$}
        & \textbf{MuSR} & \textbf{$\Delta$}
        & \textbf{BBH} & \textbf{$\Delta$}
        & \\
        \midrule
        \textit{${Qwen2.5(14B)_T}$}
        & \textit{$98.53_{\pm 0.32}$} & --
        & \textit{$84.91_{\pm 0.62}$} & --
        & \textit{$56.48_{\pm 0.83}$} & --
        & \textit{$80.10_{\pm 0.13}$} & --
        & \textit{86.80} & --
        & \textit{75.55} & --
        & \textit{41.01} & --
        & \textit{51.71} & --
        & -- \\
        \midrule
        ${Qwen2.5(1.5B)_{S}}$
        & $96.73_{\pm 0.70}$ & \deltaz
        & $56.71_{\pm 1.03}$ & \deltaz
        & $43.08_{\pm 1.41}$ & \deltaz
        & $\mathbf{47.09}_{\pm 17.17}$ & \deltaz
        & 89.50 & \deltaz
        & 53.57 & \deltaz
        & 36.90 & \deltaz
        & 36.19 & \deltaz
        & 2.4 \\
        \midrule
        Reverse KL
        & $97.37_{\pm 0.25}$ & \deltag{0.64}{20}
        & $56.30_{\pm 1.43}$ & \deltar{0.41}{18}
        & $44.66_{\pm 0.68}$ & \deltag{1.58}{25}
        & $44.52_{\pm 0.40}$ & \deltar{2.57}{30}
        & 80.00 & \deltar{9.50}{80}
        & 52.17 & \deltar{1.40}{24}
        & 35.85 & \deltar{1.05}{22}
        & 32.70 & \deltar{3.49}{38}
        & 3.0 \\
        TSD-KD
        & $97.20_{\pm 0.56}$ & \deltag{0.47}{18}
        & $52.55_{\pm 2.01}$ & \deltar{4.16}{45}
        & $42.46_{\pm 2.20}$ & \deltar{0.62}{18}
        & $37.08_{\pm 1.55}$ & \deltar{10.01}{80}
        & $\mathbf{93.60}$ & \deltag{4.10}{45}
        & 53.44 & \deltar{0.13}{18}
        & $\mathbf{39.81}$ & \deltag{2.91}{35}
        & $\mathbf{37.96}$ & \deltag{1.77}{28}
        & 2.6 \\
        \oursrow
        \textbf{RG-OPD (ours)}
        & $\mathbf{97.80}_{\pm 0.20}$ & \deltag{1.07}{24}
        & $\mathbf{59.46}_{\pm 0.48}$ & \deltag{2.75}{35}
        & $\mathbf{44.89}_{\pm 1.45}$ & \deltag{1.81}{28}
        & $44.04_{\pm 0.31}$ & \deltar{3.05}{35}
        & 79.00 & \deltar{10.50}{80}
        & $\mathbf{53.85}$ & \deltag{0.28}{18}
        & 37.83 & \deltag{0.93}{22}
        & 34.40 & \deltar{1.79}{28}
        & \textbf{2.0} \\
        \bottomrule
    \end{tabular}%
    }
\end{table*}
\end{document}